\documentclass[a4paper,twoside]{article}

\usepackage{epsfig}
\usepackage{graphicx}
\usepackage{tabularx}
\usepackage{subcaption}
\usepackage{calc}
\usepackage{amssymb}
\usepackage{amstext}
\usepackage{amsmath}
\usepackage{amsthm}
\usepackage{multicol}
\usepackage{pslatex}
\usepackage{apalike}
\usepackage{algorithm2e}
\usepackage[bottom]{footmisc}
\usepackage{SCITEPRESS}     

\begin{document}

\title{CRAFT: Clustered Regression for Adaptive Filtering of Training data}

\author{\authorname{
Parthasarathi Panda\sup{1}\orcidAuthor{0009-0000-4520-1840},
Asheswari Swain\sup{2}\orcidAuthor{0009-0002-2692-2487}, 
Subhrakanta Panda \sup{3} \orcidAuthor{0000-0003-4768-772X}}
\affiliation{\sup{1}Google, USA}
\affiliation{\sup{2, 3}Department of Computer Science \& Information Systems, BITS Pilani, Hyderabad, India}
\email{foopanda@google.com, 2015hw68129@wilp.bits-pilani.ac.in, spanda@hyderabad.bits-pilani.ac.in}
}


\abstract{
Selecting a small, high-quality subset from a large corpus for fine-tuning is increasingly important as corpora grow to tens of millions of datapoints, making full fine-tuning expensive and often unnecessary. We propose CRAFT (Clustered Regression for Adaptive Filtering of Training data), a vectorization-agnostic selection method for training sequence-to-sequence models. CRAFT decomposes the joint source-target distribution and performs a two-stage selection: (i) match the validation source distribution through proportional budget allocation across $k$-means clusters, and (ii) within each source cluster, select training pairs whose target embeddings minimize a conditional expected distance derived from the validation target distribution. We prove that proportional cluster allocation bounds the continuous KL divergence between selected and validation distributions, with the residual controlled by cluster diameters. We evaluate CRAFT on English-Hindi translation by selecting training data from 33 million NLLB sentence pairs and fine-tuning mBART via LoRA. CRAFT achieves 43.34 BLEU, outperforming TSDS (41.21) by 2.13 points on the same candidate pool and encoder while completing selection over 40$\times$ faster. With TF-IDF vectorization, the entire pipeline completes in under one minute on CPU. TAROT achieves 45.61 BLEU, but CRAFT completes selection in 26.86 seconds versus TAROT's 75.6 seconds, a 2.8$\times$ speedup.}

\onecolumn \maketitle \normalsize \setcounter{footnote}{0} \vfill

\section{\uppercase{Introduction}}
\label{sec:introduction}

\noindent The effectiveness of fine-tuned neural machine translation (NMT) models critically depends on the quality and relevance of their training data. As parallel corpora grow to tens of millions of sentence pairs, such as the 33 million English-Hindi pairs in the NLLB corpus \cite{fan2020englishcentricmultilingualmachinetranslation}, selecting appropriate training data has emerged as a critical challenge \cite{albalak2024survey}. Full fine-tuning on large corpora is computationally expensive and often unnecessary: a small, well-chosen subset can match or exceed the performance of training on the full dataset, provided the selection captures the right distributional properties.

Formally, given a validation dataset $V = \{(s_i, t_i)\}_{i=1}^{M}$ and a large training pool $T = \{(s_i, t_i)\}_{i=1}^{N}$, we investigate whether we can select a subset $T' \subset T$ where $|T'| = k \ll N$ such that fine-tuning on $T'$ maximizes model performance on $V$. The selected subset should satisfy two requirements: (1) the selection algorithm should complete much faster than training on the full dataset, and (2) the model trained on $T'$ should significantly outperform one trained on a randomly selected subset of the same size.

Existing methods address this problem with varying trade-offs between quality and computational cost. Lexical methods such as DSIR---Data Selection via Importance Resampling---\cite{xie2023data} are fast but fail to capture semantic structure. Gradient-based methods such as LESS---Low-rank gradiEnt Similarity Search---\cite{xia2024less} and TSDS---Task Specific Data Selection---\cite{liu2024tsds} achieve strong performance but require expensive encoder inference or optimal transport solves over the full candidate pool. TAROT---Targeted data selection framework grounded in Optimal Transport---\cite{feng2024tarot} achieves the highest model performance in our experiments but uses a greedy optimal transport procedure that is considerably slower at selection time than our method.

A key structural property of parallel data that prior methods do not exploit is the conditional relationship between source and target sentences. Methods that embed source and target jointly treat each sentence pair as a single point, losing the conditional structure $P(\text{target} \mid \text{source})$ that characterizes sentence-to-sentence tasks. We propose CRAFT (Clustered Regression for Adaptive Filtering of Training data), which explicitly models this structure by clustering source and target embeddings separately and performing a two-stage selection: (1) matching the validation source distribution through proportional cluster budget allocation, and (2) selecting training targets that minimize a conditional expected distance to the validation target distribution. This factored strategy is theoretically grounded as we prove it bounds the continuous KL divergence between selected and validation distributions. It is also practically efficient, completing selection in 26.86 seconds on a 1 million candidate pool versus TAROT's 75.6 seconds and TSDS's 18.1 minutes.

A further distinguishing property of CRAFT is that it is \emph{vectorization-agnostic}: the clustering, conditional probability estimation, and scoring depend only on distances between vectors, not on how those vectors are produced. We demonstrate this with both dense multilingual sentence embeddings and TF-IDF vectors. The TF-IDF variant selects from 1 million candidates in under one minute on a CPU with no GPU required, while achieving performance comparable to TSDS with dense embeddings (41.78 vs.\ 41.21 BLEU). The rest of the paper is structured as follows: Section~\ref{sec:background} reviews related work, Section~\ref{sec:approach2} presents the CRAFT algorithm and its theoretical justification, Section~\ref{sec:approach3} describes the TF-IDF instantiation, Section~\ref{sec:experiments} reports experimental results, and Section~\ref{sec:conclusion} concludes.

\section{\uppercase{Background and Related Work}}
\label{sec:background}

\noindent Several recent methods formalize data selection as distribution matching between a small set of representative target examples and a large candidate pool, spanning a spectrum of feature representations and selection mechanisms.

At one end of this spectrum, DSIR \cite{xie2023data} uses surface-level lexical features for importance resampling. Each candidate is featurized using hashed $n$-gram features and importance weights are computed as the ratio $w_i = \hat{p}_{\text{target}}(z_i) / \hat{q}_{\text{raw}}(z_i)$ between target and raw distributions, with candidates sampled without replacement according to these normalized weights. DSIR scales to 100 million documents in 4.5 hours and performs comparably to expert curation, but its reliance on surface-level features limits its ability to capture deeper structural or reasoning patterns.

Moving to model-aware representations, LESS \cite{xia2024less} addresses data selection using influence functions. It computes gradients with respect to LoRA parameters, projects them via random projection, and scores each candidate by its maximum cosine similarity to any target example. While LESS demonstrates that training on a selected 5\% often outperforms the full dataset, it requires expensive gradient computation over the entire candidate pool.

At the other end, TSDS \cite{liu2024tsds} formulates data selection as an optimal transport problem with explicit regularization for diversity, proposing KNN-Uniform and KNN-KDE instantiations that admit closed-form solutions via nearest-neighbor search using FAISS, Facebook AI Similarity Search \cite{johnson2019billion}. Using gradient-based encodings, TSDS outperforms LESS by an average of 1.5 F1 points at a 1\% selection ratio.

More recently, TAROT \cite{feng2024tarot} extends the optimal transport perspective by identifying two limitations of influence-based greedy heuristics: the disproportionate impact of dominant feature components in high-dimensional influence estimation, and the restrictive linear additive assumptions inherent in greedy selection. As the scale and diversity of training data grow, effective data selection has become increasingly important for both language and vision models \cite{albalak2024survey}, \cite{liu2024less}, \cite{kang2024performance}, \cite{engstrom2024dsdm}. TAROT addresses these challenges by incorporating whitened feature distance to mitigate dominant feature bias and then minimizing the optimal transport distance between the selected data and target domains. Evaluated across semantic segmentation, motion prediction, and instruction tuning, TAROT consistently outperforms prior methods including LESS and TSDS, while also enabling estimation of optimal selection ratios.

Our approach differs from these methods in two key respects. First, we operate on parallel sentence pairs, where source and target languages carry complementary distributional information. While prior methods treat each example as a single vector, we cluster source and target representations separately, capturing the conditional structure $P(\text{target cluster} \mid \text{source cluster})$ from the validation set. This discards the exact distribution matching paradigm attempted by previous work. Second, we offer approaches agnostic to vector representations, so that we can use the whole gamut of vector representations available from TF-IDF to model specific gradients.

\section{\uppercase{Proposed Approach (CRAFT)}}
\label{sec:approach2}

Let $(S, T)$ denote the random variable representing a pair of source-target vectors in some vector space. The validation set $V$ induces an empirical joint distribution $\hat{P}_V(S, T)$ that characterizes the target task. Given a selection $K \subseteq [N]$ from the training pool where $N$ is the candidate pool size , we want the selected pairs to be useful for learning the translation patterns present in $V$.

By the chain rule of probability, the joint distribution decomposes as:
\begin{equation}
P(S, T) = P(S) \cdot P(T \mid S)
\label{eq:chain}
\end{equation}

This decomposition suggests a two-stage strategy: first match the marginal source distribution $P(S)$, then, conditioned on each source type, select training targets that are close to the validation targets. For the source marginal, we perform distribution matching. For the target conditional, we minimize an expected distance loss. Specifically, we define the loss for a selected subset with conditional target distribution $P_{K}(T \mid S)$:
\begin{equation}
\begin{split}
\mathcal{L}(K) = \mathbb{E}_{S \sim P_V(S)} \Big[ \iint & \| t - t_V \| \\
& \cdot P_K(T{=}t \mid S) \\
& \cdot P_V(T{=}t_V \mid S) \, dt \, dt_V \Big]
\end{split}
\label{eq:loss}
\end{equation}

This measures the expected distance between training targets and validation targets, where the source is drawn from the validation marginal and both target distributions are conditioned on the same source. The approach is a \emph{hybrid}: distributional on the source side (matching $P(S)$), geometric on the target side (minimizing $\mathbb{E}[\|T - T_V\| \mid S]$). This factored strategy is a principle behind stratified sampling \cite{cochran1977sampling}, where the population is partitioned into strata and within-stratum selection is optimized separately. 

\begin{figure}[!htbp]
\centering
\begin{subfigure}[t]{0.48\textwidth}
    \centering
    \includegraphics[width=\textwidth]{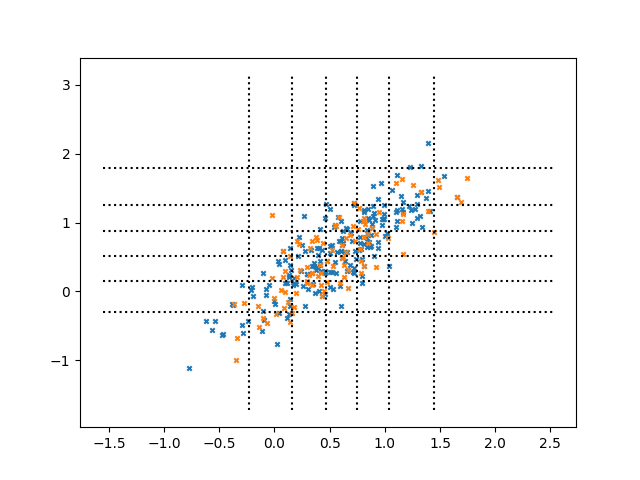}
    \caption{Distribution-matched: two samples drawn from the same joint distribution. The selected points (orange) cover the full spread of the validation points (blue).}
    \label{fig:dist_matched}
\end{subfigure}
\hfill
\begin{subfigure}[t]{0.48\textwidth}
    \centering
    \includegraphics[width=\textwidth]{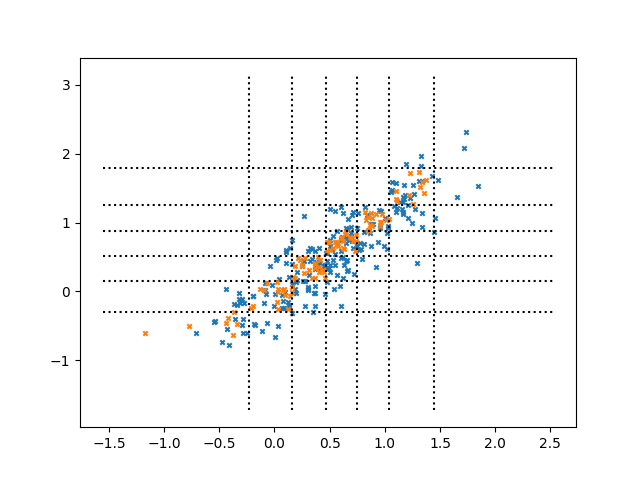}
    \caption{CRAFT's selection: selected points (orange) are concentrated along the conditional structure $P(T \mid S)$, producing a ``thinner'' subset that captures the source-target correlation.}
    \label{fig:conditional}
\end{subfigure}
\caption{Distribution Matching vs CRAFT}\label{fig:motivation}
\end{figure}

Figure \ref{fig:motivation} illustrates how this differs from direct distribution matching. The $x$-axis represents the source embedding and the $y$-axis the target embedding and the dashed lines indicate cluster boundaries. Na\"ive distribution matching selects points that cover the full joint distribution but includes many pairs far from the conditional structure. Our approach selects points along the conditional relationship, concentrating on pairs where the target is close to the expected target given the source cluster.

\subsection{Discretization via Clustering}\label{sec:discretization}

We discretize the continuous feature spaces by performing $k$-means clustering on the source and target embeddings of the validation set separately. This simplifies the continuous objective in Eq.~\ref{eq:loss} into a tractable combinatorial problem.

We define the following embedding sets:
\begin{itemize}
    \item $V_s^\dagger = \{u_i\}_{i=1}^{M}$: source vector representations of validation sentences
    \item $V_t^\dagger = \{v_i\}_{i=1}^{M}$: target vector representations of validation sentences
    \item $T_s^\dagger = \{x_i\}_{i=1}^{N}$: source vector representations of training sentences
    \item $T_t^\dagger = \{y_i\}_{i=1}^{N}$: target vector representations of training sentences
\end{itemize}

Performing $k$-means clustering on $V_s^\dagger$ and $V_t^\dagger$ separately yields source clusters $\{R_1, \ldots, R_{m_s}\}$ with centroids $\{c_1, \ldots, c_{m_s}\}$ and target clusters $\{S_1, \ldots, S_{m_t}\}$ with centroids $\{\mu_1, \ldots, \mu_{m_t}\}$. Each cluster assignment function $\pi_s: \mathbb{R}^d \to \{1, \ldots, m_s\}$ and $\pi_t: \mathbb{R}^d \to \{1, \ldots, m_t\}$ induces discrete random variables $\tilde{S} = \pi_s(S)$ and $\tilde{T} = \pi_t(T)$.

The discretized joint distribution from the validation set is:
\begin{equation}
\hat{P}_V(\tilde{S} = i, \tilde{T} = j) = \frac{|\{k \in [M] : u_k \in R_i \text{ and } v_k \in S_j\}|}{M}
\end{equation}

which factors as:
\begin{equation}
\hat{P}_V(\tilde{S} = i, \tilde{T} = j) = \hat{P}_V(\tilde{S} = i) \cdot \hat{P}_V(\tilde{T} = j \mid \tilde{S} = i)
\label{eq:factored}
\end{equation}

Let $d_{jk} = \|\mu_j - \mu_k\|$ be the distance between target cluster centroids. Under the cluster discretization, the continuous loss in Eq.~\ref{eq:loss} simplifies to:
\begin{equation}
\hat{\mathcal{L}}(K) = \sum_{i} \sum_{j, k} d_{jk} \cdot \hat{P}_V(\tilde{T} = j \mid \tilde{S} = i) \cdot q_{k|i}(K)
\label{eq:discrete_loss}
\end{equation}
where $q_{k|i}(K) = P_K(\pi_t(T) = k \mid \pi_s(S) = i)$ is the conditional target cluster distribution of the selected subset within source cluster $i$.

\subsection{Two-Stage Optimization}

Given a selection budget of $|K| = k$ training pairs, we formulate the selection as a two-stage optimization:

\textbf{Stage 1: Source marginal matching (distributional).} We allocate the budget across source clusters proportionally to the validation marginal:
\begin{equation}
k_i = \lfloor k \cdot \hat{P}_V(\tilde{S} = i) \rceil
\label{eq:budget}
\end{equation}
where $\lfloor \cdot \rceil$ denotes rounding with redistribution to ensure $\sum_i k_i = k$. This guarantees $\hat{P}_{T'}(\tilde{S} = i) = k_i / k \approx \hat{P}_V(\tilde{S} = i)$.
 
\subsubsection{Justification: $k$-means Proportional Allocation as Distribution Matching}
 
Proportional allocation across $k$-means clusters is a heuristic for matching the continuous validation distribution $P_V(S)$. We now show that it minimizes one term of a decomposition of the continuous KL divergence, and that the remaining term is controlled by the cluster diameters.
 
Suppose we have continuous density functions $p_V(s)$ and $p_T(s)$ for the validation and training distributions, and we have partitioned the feature space into $\{R_1, \ldots, R_{m_s}\}$ via $k$-means. Let $\hat{P}_V^{(m)}(i) = P_V(S \in R_i) = \int_{R_i} p_V(s)\,ds$ and $\hat{P}_{T'}^{(m)}(i) = P_{T'}(S \in R_i) = \int_{R_i} p_T(s)\,ds$ denote the discretized distributions over the partition.
 
We decompose the continuous Kullback-Leibler (KL) divergence, a measure of how close two distributions are \cite{kldivergence}, by splitting the integral across clusters and introducing discretized probabilities:
\begin{equation}
\begin{split}
D_{\text{KL}}(P_V \| P_{T'}) &= \int p_V(s)\log\frac{p_V(s)}{p_T(s)}\,ds \\
&= \sum_{i=1}^{m_s}\int_{R_i} p_V(s)\log\frac{p_V(s) / \hat{P}_V^{(m)}(i)}{p_T(s) / \hat{P}_{T'}^{(m)}(i)}\,ds \\
&\quad + \sum_{i=1}^{m_s} \hat{P}_V^{(m)}(i) \log\frac{\hat{P}_V^{(m)}(i)}{\hat{P}_{T'}^{(m)}(i)}
\end{split}
\label{eq:kl1}
\end{equation}
 
Defining $d_i = \int_{R_i} p_V(s)\log\frac{p_V(s)/\hat{P}_V^{(m)}(i)}{p_T(s)/\hat{P}_{T'}^{(m)}(i)}\,ds$, Eq.~\ref{eq:kl1} becomes:
\begin{equation}
D_{\text{KL}}(P_V \| P_{T'}) = \sum_{i=1}^{m_s} d_i + D_{\text{KL}}(\hat{P}_V^{(m)} \| \hat{P}_{T'}^{(m)})
\label{eq:kl2}
\end{equation}
 
The continuous KL divergence thus decomposes into two terms: (1) the discretized KL divergence $D_{\text{KL}}(\hat{P}_V^{(m)} \| \hat{P}_{T'}^{(m)})$ over cluster probabilities, and (2) the within-cluster terms $d_i$.
 
\textbf{Proportional allocation minimizes the discretized KL.} The proportional allocation ensures $\hat{P}_{T'}^{(m)}(i) = k_i / k \approx \hat{P}_V^{(m)}(i)$, so:
\begin{equation}
D_{\text{KL}}(\hat{P}_V^{(m)} \| \hat{P}_{T'}^{(m)}) \approx 0
\label{eq:kl_marginal}
\end{equation}
The approximation error arises only from integer rounding of $k \cdot p_i$ and vanishes as $k \to \infty$.
 
\textbf{Bounding the within-cluster terms.} Assume $p_V(s)$ and $p_T(s)$ are Lipschitz continuous. Note that the conditional densities within each cluster integrate to one:
\begin{equation}
\int_{R_i} \frac{p_V(s)}{\hat{P}_V^{(m)}(i)}\,ds = \int_{R_i} \frac{p_T(s)}{\hat{P}_{T'}^{(m)}(i)}\,ds = 1
\end{equation}
 
By the mean value theorem, there exists $s_i \in R_i$ such that $\frac{p_V(s_i)}{\hat{P}_V^{(m)}(i)} = \frac{p_T(s_i)}{\hat{P}_{T'}^{(m)}(i)}$. Define the shorthand $\ell(s) = \log\frac{p_V(s)/\hat{P}_V^{(m)}(i)}{p_T(s)/\hat{P}_{T'}^{(m)}(i)}$. Since $p_V$ and $p_T$ are Lipschitz continuous, $\ell(s)$ is also Lipschitz with some constant $L_\ell$, and $\ell(s_i) = 0$. Let $\epsilon_i = \max_{s_1, s_2 \in R_i} \|s_1 - s_2\|$ be the diameter of cluster $R_i$. Then:
\begin{equation}
\ell(s) = \ell(s) - \ell(s_i) \leq L_\ell \|s - s_i\| \leq L_\ell \epsilon_i
\end{equation}
 
This bounds each within-cluster term:
\begin{equation}
d_i = \int_{R_i} p_V(s) \, \ell(s) \, ds \leq \int_{R_i} p_V(s) \, L_\ell \epsilon_i \, ds = \hat{P}_V^{(m)}(i) \, L_\ell \epsilon_i
\end{equation}
 
Substituting into Eq.~\ref{eq:kl2}:
\begin{equation}
D_{\text{KL}}(P_V \| P_{T'}) \leq \sum_{i=1}^{m_s} \hat{P}_V^{(m)}(i) \, L_\ell \epsilon_i + D_{\text{KL}}(\hat{P}_V^{(m)} \| \hat{P}_{T'}^{(m)})
\label{eq:kl_bound}
\end{equation}
 
The first term $\sum_i \hat{P}_V^{(m)}(i) L_\ell \epsilon_i$ is the probability-weighted average cluster diameter, scaled by the Lipschitz constant. This is controlled by the $k$-means objective, which minimizes within-cluster variance $\sum_i \sum_{u \in R_i} \|u - c_i\|^2$. While $k$-means does not directly minimize the diameter $\epsilon_i$, variance and diameter are closely related: small within-cluster variance implies that most points are close to the centroid, and both quantities shrink as $m_s$ increases. The second term is driven to zero by proportional allocation (Eq.~\ref{eq:kl_marginal}).
 
Thus, $k$-means proportional allocation minimizes the continuous KL divergence up to a residual that depends on cluster diameters, which are themselves minimized by the $k$-means objective. Increasing $m_s$ shrinks each $\epsilon_i$, tightening the bound at the cost of noisier cluster proportion estimates from finite validation data.

\textbf{Stage 2: Conditional target distance minimization (geometric).} 
Eq.~\ref{eq:discrete_loss} given in Section~\ref{sec:discretization} is a linear expression in $q_{k|i}(K)$. Thus to minimize it, we select $k_i$ candidates from the bucket $B_i = \{l \in [N] : \pi_s(x_l) = i\}$ with the smallest $c(l, i)$. Here $c(l, i)$ is the per candidate cost computed as:
\begin{equation}
c(l, i) = \sum_{j=1}^{m_t} p(j \mid i) \cdot d_{j, \pi_t(y_l)}
\label{eq:candidate_cost}
\end{equation}

We select the $k_i$ candidates from $B_i$ with the smallest costs.

\subsection{Design Choice: Cluster Centroids vs.\ Raw Embeddings}

A natural alternative would be to score using the raw target embedding $y_l$ directly: $c(l, i) = \sum_j p(j|i) \|\mu_j - y_l\|$. We deliberately use the cluster centroid $\mu_{\pi_t(y_l)}$ instead of $y_l$, for two reasons:

\begin{enumerate}
\item \textbf{Regularization against overfitting.} Using the raw embedding would select candidates whose targets are nearest to the conditional centroid $\bar{\mu}_i = \sum_j p(j|i) \mu_j$ in embedding space. This risks \emph{overfitting}: the selected training set would cluster tightly around a single point in target space, and a model trained on such data may fail to generalize beyond the narrow region covered by the validation set. By scoring at the granularity of cluster centroids, candidates anywhere within the same target cluster receive the same score, preserving diversity within each cluster.

\item \textbf{Robustness to metric mismatch.} The Euclidean distance $\|y - y'\|$ in embedding space is a proxy for the true loss function used during model training (e.g., cross-entropy over token sequences for NMT). Fine-grained distance comparisons between individual embeddings may not correlate well with the actual training loss. Operating at the cluster level provides a coarser but more robust signal: two candidates in the same target cluster are treated as equivalent regardless of their exact positions, which is appropriate when the embedding metric is only an approximation.
\end{enumerate}

\textbf{Structural consequence.} The cost $c(l, i)$ depends on candidate $l$ only through its target cluster assignment $\pi_t(y_l)$, not its exact position. This partitions candidates in each source cluster $B_i$ into $m_t$ groups, with all candidates in the same group receiving the same cost. Selection proceeds by ranking target clusters within each source cluster by their conditional alignment, and filling the budget from the best-aligned clusters first.

\subsection{Algorithm}

The algorithm \ref{alg:approach2} summarizes the approach. It begins with an initialization step that performs $k$-means on the validation dataset. This is followed by stage-$1$ where buckets and costs for each candidate data point are identified. In stage-$2$ we identify the target counts from each bucket and pick accordingly based on the costs. \\
\textbf{Complexity.} Let $E$ denote the cost of vectorizing one sentence. The vectorization step is $O((M+N) \cdot E)$. Clustering is $O(M \cdot m \cdot d)$ for $k$-means iterations. Training point assignment is $O(N \cdot (m_s + m_t) \cdot d)$. The inter-centroid distance matrix $D$ is $O(m_t^2 \cdot d)$. Scoring is $O(N \cdot m_t)$. The min-$k_a$ selection within each source cluster can be performed via Quickselect in expected $O(|B_a|)$ time, or via a min-heap in $O(|B_a| \log k_a)$ worst case; summed over all clusters this gives expected $O(N)$ or $O(N \log k)$ respectively. The dominant cost is vectorization when using LLM-based encoders ($O(N \cdot E)$), or cluster assignment ($O(N \cdot m \cdot d)$) when using TF-IDF.

\begin{algorithm}[!h]
 \KwIn{Validation set $V = \{(s_i, t_i)\}_{i=1}^M$, Training set $T = \{(s_i, t_i)\}_{i=1}^N$, target size $k$, source clusters $m_s$, target clusters $m_t$}
 \KwOut{Selected subset $T' \subset T$, $|T'| = k$}
 
 \textbf{/* Vectorization */}\;
 Vectorize source sentences: $V_s^\dagger = \{u_i\}$, $T_s^\dagger = \{x_i\}$\;
 Vectorize target sentences: $V_t^\dagger = \{v_i\}$, $T_t^\dagger = \{y_i\}$\;
 
 \textbf{/* Clustering on validation set */}\;
 Run $k$-means on $V_s^\dagger$ with $m_s$ clusters $\to$ centroids $\{c_1, \ldots, c_{m_s}\}$\;
 Run $k$-means on $V_t^\dagger$ with $m_t$ clusters $\to$ centroids $\{\mu_1, \ldots, \mu_{m_t}\}$\;
 
 \textbf{/* Compute validation distribution */}\;
 \ForEach{validation pair $(u_i, v_i)$}{
   $\pi_s(i) \leftarrow \arg\min_a \|u_i - c_a\|$\;
   $\pi_t(i) \leftarrow \arg\min_b \|v_i - \mu_b\|$\;
 }
 $p_a \leftarrow |\{i : \pi_s(i) = a\}| / M$ for each $a \in [m_s]$\;
 $p(b \mid a) \leftarrow |\{i : \pi_s(i) = a \text{ and } \pi_t(i) = b\}| / |\{i : \pi_s(i) = a\}|$ for each $a, b$\;
 
 \textbf{/* Assign training points to clusters */}\;
 \ForEach{training pair $(x_l, y_l)$}{
   $\pi_s(l) \leftarrow \arg\min_a \|x_l - c_a\|$\;
   $\pi_t(l) \leftarrow \arg\min_b \|y_l - \mu_b\|$\;
 }
 
 \textbf{/* Stage 1: Budget allocation (source matching) */}\;
 $k_a \leftarrow \lfloor k \cdot p_a \rceil$ for each $a \in [m_s]$, adjusted so $\sum_a k_a = k$\;
 $B_a \leftarrow \{l : \pi_s(l) = a\}$ for each $a$\;
 
 \textbf{/* Precompute inter-centroid costs */}\;
 $C_{b,b'} \leftarrow  \sum_{a=1}^{m_t} p(a \mid b) \cdot \|\mu_a - \mu_{b'}\|$
for each $b, b' \in [m_t]$\;
 
 \textbf{/* Stage 2: Score and select within each source cluster */}\;
 $T' \leftarrow \emptyset$\;
 \ForEach{source cluster $a \in [m_s]$}{
   \ForEach{$l \in B_a$}{
     $c(l) \leftarrow C_{a,\pi_t(l)}$\;
   }
   $T' \leftarrow T' \cup \text{MinK}(B_a, c, k_a)$\;
 }
 \Return $T'$\;
\caption{CRAFT.}\label{alg:approach2}
\end{algorithm}

\subsection{Comparison with Related Approaches}

\begin{enumerate}
\item \textbf{Relation to TSDS.} TSDS \cite{liu2024tsds} minimizes a global transport cost $\sum_{i,j} \gamma_{ij} d_{ij}$ between query examples and candidates with a diversity regularizer. Our approach decomposes this via $P(S,T) = P(S) \cdot P(T|S)$: Stage~1 (source matching) is analogous to the transport constraint $\sum_j \gamma_{ij} = 1/M$, while Stage~2 (target scoring) is analogous to the transport cost, but operates within each source stratum independently. TSDS requires a separate KDE-based regularizer to prevent over-concentration on near-duplicates; our use of cluster-level scoring provides implicit regularization by treating all candidates within a target cluster as equivalent.

\item \textbf{Relation to DSIR.} DSIR \cite{xie2023data} performs global distribution matching via importance resampling, minimizing KL divergence in $n$-gram feature space. Our approach also matches the source distribution (Stage~1), but replaces global distributional matching on the target with a conditional geometric objective (Stage~2). Both methods avoid fine-grained embedding comparisons, DSIR through hashed $n$-gram bucketing, and our method through cluster-level scoring, which contributes to robustness.

\item \textbf{Relation to TAROT.} TAROT \cite{feng2024tarot} minimizes the optimal transport distance between selected data and the target domain, using whitened feature distances to mitigate dominant component bias. While TAROT achieves the highest translation quality in our experiments, it requires solving an optimal transport problem at selection time. CRAFT instead replaces this with a closed-form cluster-level scoring step, achieving selection in 26.86 seconds versus TAROT's 75.6 seconds on the 1M pool, a 2.8$\times$ speedup, while remaining within 2.27 BLEU points of TAROT.

\item \textbf{Parallel structure.} Methods that embed the source and target jointly (TAROT, TSDS) do not explicitly capture the conditional relationship $P(T \mid S)$. By clustering source and target independently, we model which target types correspond to which source types in the validation set. The cluster-level scoring naturally captures this structure: for each source cluster, target clusters are ranked by their conditional alignment under the validation distribution, and candidates from higher-ranked target clusters are selected first.
\end{enumerate}

\section{\uppercase{TF-IDF Based Vectorization}}
\label{sec:approach3}

\noindent A key property of the selection algorithm is that it is \emph{vectorization-agnostic}: the clustering, conditional probability estimation, and scoring depend only on distances between vectors, not on how those vectors were produced. We demonstrate this in Section~\ref{sec:experiments} by using TF-IDF, Term Frequency-Inverse Document Frequency \cite{sparckjones1972statistical} to vectorize the sentences alongside LLM-based embeddings while keeping the selection algorithm (Algorithm \ref{alg:approach2}) entirely unchanged.

This substitution provides a significant speedup: vectorizing 33,193,629 English-Hindi sentence pairs from NLLB takes approximately 265.3 minutes using LLM embeddings on an A100 GPU, whereas TF-IDF fitting and transformation of the same data takes only 16.32 minutes on a CPU, a speedup factor of approximately 16$\times$. The TF-IDF representation captures lexical similarity rather than semantic similarity, offering a different trade-off: faster computation at the cost of missing paraphrases and synonyms that dense embeddings would capture.


\section{\uppercase{Experiments}}
\label{sec:experiments}

\begin{figure}[!htbp]
  \centering
   {\epsfig{file = 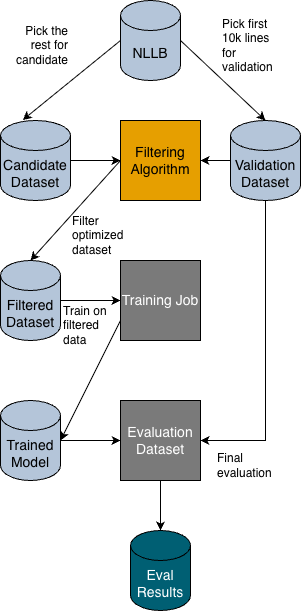, width = 5.5cm}}
  \caption{Data flow for the experiment.}
  \label{fig:example1}
 \end{figure}

We evaluate CRAFT on English--Hindi machine translation 
to answer three questions: 
\begin{enumerate}
    \item Does stratified conditional selection outperform existing 
    data selection baselines on translation quality? 
    (Section~\ref{sec:experiments}.2)
    \item How does CRAFT's selection time compare against prior 
    methods on the same candidate pool? 
    (Section~\ref{sec:experiments}.3)
    \item Is the performance gain driven by the selection algorithm itself or by the choice of dense embeddings? 
    (Sections~\ref{sec:experiments}.4 and \ref{sec:experiments}.5)
\end{enumerate} 
Figure~\ref{fig:example1} shows the end-to-end pipeline used across all experiments.

Parallel corpora for machine translation are among the largest and most heterogeneous data repositories in NLP. The NLLB corpus alone contains 33 million sentence pairs across hundreds of language pairs. This scale makes exhaustive fine-tuning computationally prohibitive, creating a practically motivated testbed where data selection methods must demonstrably work. In addition, translation tasks have cheap, reproducible metrics that allow direct comparison across methods. Hence, we picked translation as a benchmark for evaluating selection quality.

Accordingly, mBART-50 is a suitable model to train with because it is pre-trained on multiple languages with good performance including low-resource languages \cite{tranicart}, \cite{low_resource_survey}, \cite{Smirnov2022ComparativeAO}. This isolates the effect of data selection from the confound of low-resource language learning.

\subsection{Setup}
 
We evaluate on English-Hindi translation using the NLLB parallel corpus\footnote{https://opus.nlpl.eu/datasets/NLLB} \cite{schwenk2020ccmatrixminingbillionshighquality}, \cite{fan2020englishcentricmultilingualmachinetranslation} obtained via OPUS \cite{tiedemann-2012-parallel}, containing 33,193,629 sentence pairs. The first 10,000 pairs serve as validation (query) examples for data selection, and the next 33,183,629 form the candidate pool.
 
Each selected training subset (20,000 pairs) is used to fine-tune mBART-50 \cite{mbart2021} using LoRA \cite{hu2022lora}. We report the following scores 
\begin{itemize}
    \item BLEU, Bilingual Evaluation Understudy \cite{BLEU}
    \item chrF, Character n-gram F-score \cite{chrf}
    \item METEOR, Metric for Evaluation of Translation with Explicit ORdering \cite{meteor}
\end{itemize}
These are popular metrics for scoring machine translation in a fast and low-cost manner. 

We compare CRAFT against DSIR \cite{xie2023data}, TSDS \cite{liu2024tsds} and TAROT \cite{feng2024tarot}. We use official implementations for DSIR\footnote{https://github.com/p-lambda/dsir} and TSDS\footnote{https://github.com/ZifanL/TSDS}. TAROT and TSDS involve vectorizations by embedding the source and target sentences separately using \texttt{paraphrase-multilingual-MiniLM-L12-v2} \cite{paraphrase} and concatenating them to produce a single vector per datapoint. For a fair comparison between CRAFT, TSDS, and TAROT, all three methods use the same candidate pool size and the same sentence encoder unless otherwise noted. 

Table \ref{tab:hyperparam} describes the hyperparameters used for the fine-tuning. LoRA rank and $\alpha$ follow the standard defaults from \cite{hu2022lora}, with rank 16 providing sufficient capacity for domain adaptation without over-parameterizing the 600M-parameter mBART-50 model. Dropout of 0.05 applies light regularization appropriate for fine-tuning on small selected subsets. Batch size, weight decay, and warmup ratio (5\% of total steps) follow standard practice for sequence-to-sequence fine-tuning \cite{mbart2021}. The learning rate of 1.6e-3 was selected by grid search on the validation set; higher values produced training instability and lower values showed no improvement in BLEU.

\begin{table}[!htbp]
\caption{Hyperparameters used for training.}\label{tab:hyperparam} \centering
\begin{tabular}{|l|c|}
  \hline
  \textbf{Parameter} & \textbf{Value} \\
  \hline
  LoRA rank & 16 \\
  \hline\textbf{}
  $\alpha$ (LoRA) & 32  \\
  \hline
  LoRA dropout & 0.05 \\
  \hline
  Effective Batch Size & 64 \\
  \hline
  Training Steps & 939  \\
  \hline
  Warmup Steps & 46 \\
  \hline
  Weight Decay & 0.01 \\
  \hline
  Learning Rate & 1.6e-3 \\
  \hline
 
\end{tabular}
\end{table}
 
\subsection{Results}
 
\begin{table}[!htbp]
\caption{Translation quality (20k selected pairs). CRAFT, TSDS, and TAROT on 1M candidates use the same pool and encoder for a fair comparison.} \label{tab:main} \centering
\resizebox{\columnwidth}{!}{
\begin{tabular}{|l|c|c|c|c|}
  \hline
  \textbf{Method}  & \textbf{Pool} & \textbf{BLEU} & \textbf{METEOR} & \textbf{CHRF} \\
  \hline
  Random (Set 1)  & 33M & 30.23 & 0.427 & 47.33 \\
  \hline
  Random (Set 2)  & 33M & 29.88 & 0.4296 & 47.439 \\
  \hline
  DSIR  & 33M & 27.29 & 0.4619 & 49.52 \\
  \hline
  TSDS & 1M & 41.21 & 0.5097 & 57.65 \\
  \hline
  CRAFT(TF-IDF) & 1M & 41.78 & 0.4916 & 55.77 \\
  \hline
  CRAFT(embeddings) & 1M & 43.34 & 0.5175 & 58.31 \\
  \hline
  TAROT & 1M & \textbf{45.61} & \textbf{0.5258} & \textbf{59.43} \\
  \hline
  CRAFT(TF-IDF) & 33M & 41.10 & 0.486 & 54.89 \\
  \hline
  CRAFT(embeddings) & 33M & 43.19 & 0.5149 & 58.22 \\
  \hline
\end{tabular}}
\end{table}
 
Table \ref{tab:main} summarizes the performance of models trained on datasets filtered by the various approaches. On the matched 1M candidate pool, CRAFT with embeddings achieves 43.34 BLEU, outperforming TSDS (41.21) by 2.13 points. TAROT achieves the highest BLEU score of 45.61. CRAFT with TF-IDF on the same 1M pool achieves 41.78 BLEU which is comparable to TSDS with dense embeddings while not requiring a GPU for vectorization.
 
DSIR performs below random selection ($\sim$30.00 BLEU), likely because hashed $n$-gram features cannot capture the cross-lingual structure of EN-HI parallel data. This confirms that surface-level lexical features are insufficient for NMT data selection.
 
Scaling CRAFT to the full 33M pool yields 43.19 BLEU with embeddings and 41.10 with TF-IDF. The marginal difference between 1M and 33M pools (43.34 vs.\ 43.19) suggests that the nearest neighbors of the validation examples are concentrated within the first million most-similar candidates, and additional candidates contribute diminishing returns.
 
\subsection{Selection Efficiency}
 
\begin{table}[!htbp]
\caption{Selection time comparison. Vectorization and selection times are reported separately where applicable.}\label{tab:timing} \centering
\resizebox{\columnwidth}{!}{
\begin{tabular}{|l|c|c|c|c|}
  \hline
  \textbf{Method} & \textbf{Pool} & \textbf{Vectorization} & \textbf{Vec.\ Time} & \textbf{Select Time} \\
  \hline
  DSIR & 33M & N/A & N/A & 34.7 min \\
  \hline
  TSDS & 1M & Embedding & 8.01 min & 18.1 min \\
  \hline
  TAROT & 1M & Embedding & 8.01 min & 75.6 sec \\
  \hline
  CRAFT & 1M & Embedding & 8.01 min & \textbf{26.86 sec} \\
  \hline
  CRAFT & 1M & TF-IDF & \textbf{46.3 sec} & \textbf{10.38 sec} \\
  \hline
  CRAFT & 33M & TF-IDF & 16.32 min & 94.63 sec \\
  \hline
  CRAFT & 33M & Embedding & 265.3 min & 225.8 sec \\
  \hline
\end{tabular}}
\end{table}
 
Table \ref{tab:timing} summarizes the speed of the various approaches for filtering the dataset. Among methods operating on the same 1M candidate pool with dense embeddings (8.01 minutes vectorization), CRAFT completes selection in 26.86 seconds, a 2.8$\times$ speedup over TAROT and a 40$\times$ speedup over TSDS. Although TAROT achieves a higher BLEU (45.61 vs.\ 43.34), CRAFT offers a compelling speed-quality trade-off, particularly in settings requiring rapid iteration or large-scale deployment.
 
With TF-IDF vectorization, CRAFT selects from 1M candidates in less than 57 seconds total (vectorization + selection), making it practical for rapid iteration without any GPU requirement. Even at full 33M scale with dense embeddings, the total pipeline (265.3 min vectorization + 225.8 sec selection) is dominated by the vectorization step, not the selection algorithm itself.

\subsection{Ablation: Design Choices}
\begin{table}[!htbp]
\caption{Ablation study on 20k selected pairs.}\label{tab:ablation} \centering{
\begin{tabular}{|l|c|}
  \hline
  \textbf{Variant} & \textbf{BLEU} \\
  \hline
  Joint clustering, random selection & 32.69 \\
  \hline
  Separate clustering + scoring (CRAFT) & 43.34  \\
  \hline
  
\end{tabular}}
\end{table}
 
For ablation we ran $k$-means clustering on the concatenation of source and target embeddings and performed the bucket matching with randomly selected entries from each cluster without any preference, essentially just attempting distribution matching. Doing so resulted in a drop in the BLEU score from 43.34 to 32.69 (Table \ref{tab:ablation}), barely above random ($\sim$ 30). This confirms that both key innovations---separate source-target clustering and conditional scoring---are essential. 
 
\subsection{Vectorization Agnosticism}
 
CRAFT achieves strong results across vectorization methods, confirming that the selection algorithm is the primary driver of performance. On the full 33M pool, TF-IDF achieves 41.10 BLEU (95\% of the embedding result) with a 16$\times$ speedup in vectorization time (16.32 min vs.\ 265.3 min). On the 1M pool, TF-IDF (41.78) is comparable to TSDS with dense embeddings (41.21), demonstrating that CRAFT's stratified conditional scoring compensates for the less expressive features.
 
\section{\uppercase{Conclusion}}
\label{sec:conclusion}
We presented CRAFT, a data selection method for sentence-to-sentence datasets that decomposes the source-target joint distribution via the chain rule $P(S,T) = P(S) \cdot P(T|S)$ and performs selection in two stages: source marginal matching through proportional cluster allocation, and conditional target selection by minimizing an expected distance objective derived from the validation distribution. We proved that proportional allocation over $k$-means clusters bounds the continuous KL divergence between selected and validation source distributions, with the residual controlled by cluster diameters.
 
CRAFT is vectorization-agnostic; the same algorithm achieves strong results with both dense sentence embeddings and TF-IDF features, decoupling the selection mechanism from the choice of representation. On English-Hindi translation with 33 million NLLB pairs, CRAFT outperforms TSDS by 2.13 points on a matched candidate pool with the same encoder, while completing selection more than 20 times faster. While TAROT achieves the highest BLEU score (45.61 vs.\ 43.34), CRAFT completes selection in 26.86 seconds compared to TAROT's 75.6 seconds, a 2.8$\times$ speedup, making it the preferred method in latency-sensitive settings. The use of cluster-level scoring rather than raw embedding distances provides implicit regularization against overfitting to the validation set and robustness to metric mismatch between the embedding space and the actual training loss.
 
A few limitations of the experimental setup can be highlighted here. The datasets and tasks on which we have run these experiments are limited in scope and can be expanded to tasks beyond translation. Experiments with non-textual datasets like images are also warranted. We aim to overcome these limitations in our future work.

\section*{\uppercase{Acknowledgements}}
The authors acknowledge the use of Claude AI \cite{anthropic2024claude},
developed by Anthropic, as a writing assistance tool to help polish and
refine the prose of this manuscript.

\bibliographystyle{apalike}
{\small
\bibliography{reference}}

@inproceedings{schwenk2020ccmatrixminingbillionshighquality,
  title={CCMatrix: Mining Billions of High-Quality Parallel Sentences on the Web},
  author={Holger Schwenk and Guillaume Wenzek and Sergey Edunov and Edouard Grave and Armand Joulin},
  booktitle={Annual Meeting of the Association for Computational Linguistics},
  year={2019},
  url={https://api.semanticscholar.org/CorpusID:207863306}
}

@article{fan2020englishcentricmultilingualmachinetranslation,
  author  = {Angela Fan and Shruti Bhosale and Holger Schwenk and Zhiyi Ma and Ahmed El-Kishky and Siddharth Goyal and Mandeep Baines and Onur Celebi and Guillaume Wenzek and Vishrav Chaudhary and Naman Goyal and Tom Birch and Vitaliy Liptchinsky and Sergey Edunov and Michael Auli and Armand Joulin},
  title   = {Beyond English-Centric Multilingual Machine Translation},
  journal = {Journal of Machine Learning Research},
  year    = {2021},
  volume  = {22},
  number  = {107},
  pages   = {1--48},
  url     = {http://jmlr.org/papers/v22/20-1307.html}
}

@inproceedings{tiedemann-2012-parallel,
    title = "Parallel Data, Tools and Interfaces in {OPUS}",
    author = {Tiedemann, J{\"o}rg},
    editor = "Calzolari, Nicoletta  and
      Choukri, Khalid  and
      Declerck, Thierry  and
      Do{\u{g}}an, Mehmet U{\u{g}}ur  and
      Maegaard, Bente  and
      Mariani, Joseph  and
      Moreno, Asuncion  and
      Odijk, Jan  and
      Piperidis, Stelios",
    booktitle = "Proceedings of the Eighth International Conference on Language Resources and Evaluation ({LREC}'12)",
    month = may,
    year = "2012",
    address = "Istanbul, Turkey",
    publisher = "European Language Resources Association (ELRA)",
    url = "https://aclanthology.org/L12-1246/",
    pages = "2214--2218"
}

@inproceedings{
xie2023data,
title={Data Selection for Language Models via Importance Resampling},
author={Sang Michael Xie and Shibani Santurkar and Tengyu Ma and Percy Liang},
booktitle={Thirty-seventh Conference on Neural Information Processing Systems},
year={2023},
url={https://openreview.net/forum?id=uPSQv0leAu}
}

@inproceedings{xia2024less,
author = {Xia, Mengzhou and Malladi, Sadhika and Gururangan, Suchin and Arora, Sanjeev and Chen, Danqi},
title = {LESS: selecting influential data for targeted instruction tuning},
year = {2024},
publisher = {JMLR.org},
booktitle = {Proceedings of the 41st International Conference on Machine Learning},
articleno = {2221},
numpages = {29},
location = {Vienna, Austria},
series = {ICML'24}
}

@inproceedings{
liu2024tsds,
title={{TSDS}: Data Selection for Task-Specific Model Finetuning},
author={Zifan Liu and Amin Karbasi and Theodoros Rekatsinas},
booktitle={The Thirty-eighth Annual Conference on Neural Information Processing Systems},
year={2024},
url={https://openreview.net/forum?id=wjbTHLUSzU}
}

@article{johnson2019billion,
    author={Johnson, Jeff and Douze, Matthijs and Jégou, Hervé},
  journal={IEEE Transactions on Big Data}, 
  title={Billion-Scale Similarity Search with GPUs}, 
  year={2021},
  volume={7},
  number={3},
  pages={535-547},
  doi={10.1109/TBDATA.2019.2921572}
}

@inproceedings{
feng2024tarot,
title={{TAROT}: Targeted Data Selection via Optimal Transport},
author={Lan Feng and Fan Nie and Yuejiang Liu and Alexandre Alahi},
booktitle={Forty-second International Conference on Machine Learning},
year={2025},
url={https://openreview.net/forum?id=EznrK7QWgK}
}

@book{cochran1977sampling,
  title={Sampling Techniques},
  author={Cochran, William G.},
  edition={3rd},
  year={1977},
  publisher={John Wiley \& Sons},
  pages={65--110},
}

@inproceedings{paraphrase,
  title={Sentence-bert: Sentence embeddings using siamese bert-networks},
  author={Reimers, Nils and Gurevych, Iryna},
  booktitle={Proceedings of the 2019 conference on empirical methods in natural language processing and the 9th international joint conference on natural language processing (EMNLP-IJCNLP)},
  pages={3982--3992},
  year={2019}
}

@inproceedings{mbart2021,
    title = "Multilingual Translation from Denoising Pre-Training",
    author = "Tang, Yuqing  and
      Tran, Chau  and
      Li, Xian  and
      Chen, Peng-Jen  and
      Goyal, Naman  and
      Chaudhary, Vishrav  and
      Gu, Jiatao  and
      Fan, Angela",
    editor = "Zong, Chengqing  and
      Xia, Fei  and
      Li, Wenjie  and
      Navigli, Roberto",
    booktitle = "Findings of the Association for Computational Linguistics: ACL-IJCNLP 2021",
    month = aug,
    year = "2021",
    address = "Online",
    publisher = "Association for Computational Linguistics",
    url = "https://aclanthology.org/2021.findings-acl.304/",
    doi = "10.18653/v1/2021.findings-acl.304",
    pages = "3450--3466"
}

@article{albalak2024survey,
  title={A survey on data selection for language models},
  author={Albalak, Alon and Elazar, Yanai and Xie, Sang Michael and Longpre, Shayne and Lambert, Nathan and Wang, Xinyi and Muennighoff, Niklas and Hou, Bairu and Pan, Liangming and Jeong, Haewon and others},
  journal={arXiv preprint arXiv:2402.16827},
  year={2024}
}

@inproceedings{liu2024less,
author = {Liu, Zikang and Zhou, Kun and Zhao, Wayne Xin and Gao, Dawei and Li, Yaliang and Wen, Ji-Rong},
title = {Less is More: High-value Data Selection for Visual Instruction Tuning},
year = {2025},
isbn = {9798400720352},
publisher = {Association for Computing Machinery},
address = {New York, NY, USA},
url = {https://doi.org/10.1145/3746027.3755160},
doi = {10.1145/3746027.3755160},
booktitle = {Proceedings of the 33rd ACM International Conference on Multimedia},
pages = {3712–3721},
numpages = {10},
location = {Dublin, Ireland},
series = {MM '25}
}

@article{kang2024performance,
  title={Performance scaling via optimal transport: Enabling data selection from partially revealed sources},
  author={Kang, Feilong and Just, Huzaifa A and Sahu, Anit Kumar and Jia, Ruoxi},
  journal={Advances in Neural Information Processing Systems},
  volume={36},
  year={2024}
}

@inproceedings{engstrom2024dsdm,
author = {Engstrom, Logan and Feldmann, Axel and Madry, Aleksander},
title = {DSDM: model-aware dataset selection with datamodels},
year = {2024},
publisher = {JMLR.org},
booktitle = {Proceedings of the 41st International Conference on Machine Learning},
articleno = {498},
numpages = {36},
location = {Vienna, Austria},
series = {ICML'24}
}

@inproceedings{
hu2022lora,
title={Lo{RA}: Low-Rank Adaptation of Large Language Models},
author={Edward J Hu and yelong shen and Phillip Wallis and Zeyuan Allen-Zhu and Yuanzhi Li and Shean Wang and Lu Wang and Weizhu Chen},
booktitle={International Conference on Learning Representations},
year={2022},
url={https://openreview.net/forum?id=nZeVKeeFYf9}
}

@inproceedings{bleu,
author = {Papineni, Kishore and Roukos, Salim and Ward, Todd and Zhu, Wei-Jing},
title = {BLEU: a method for automatic evaluation of machine translation},
year = {2002},
publisher = {Association for Computational Linguistics},
address = {USA},
url = {https://doi.org/10.3115/1073083.1073135},
doi = {10.3115/1073083.1073135},
booktitle = {Proceedings of the 40th Annual Meeting on Association for Computational Linguistics},
pages = {311–318},
numpages = {8},
location = {Philadelphia, Pennsylvania},
series = {ACL '02}
}

@inproceedings{chrf,
    title = "chr{F}: character n-gram {F}-score for automatic {MT} evaluation",
    author = "Popovi{\'c}, Maja",
    editor = "Bojar, Ond{\v{r}}ej  and
      Chatterjee, Rajan  and
      Federmann, Christian  and
      Haddow, Barry  and
      Hokamp, Chris  and
      Huck, Matthias  and
      Logacheva, Varvara  and
      Pecina, Pavel",
    booktitle = "Proceedings of the Tenth Workshop on Statistical Machine Translation",
    month = sep,
    year = "2015",
    address = "Lisbon, Portugal",
    publisher = "Association for Computational Linguistics",
    url = "https://aclanthology.org/W15-3049/",
    doi = "10.18653/v1/W15-3049",
    pages = "392--395"
}

@inproceedings{meteor,
    title = "{METEOR}: An Automatic Metric for {MT} Evaluation with Improved Correlation with Human Judgments",
    author = "Banerjee, Satanjeev  and
      Lavie, Alon",
    editor = "Goldstein, Jade  and
      Lavie, Alon  and
      Lin, Chin-Yew  and
      Voss, Clare",
    booktitle = "Proceedings of the {ACL} Workshop on Intrinsic and Extrinsic Evaluation Measures for Machine Translation and/or Summarization",
    month = jun,
    year = "2005",
    address = "Ann Arbor, Michigan",
    publisher = "Association for Computational Linguistics",
    url = "https://aclanthology.org/W05-0909/",
    pages = "65--72"
}

@misc{anthropic2024claude,
  author       = {Anthropic},
  title        = {Claude: A Family of Large Language Models},
  year         = {2024},
  howpublished = {Anthropic},
  url          = {https://www.anthropic.com},
  note         = {Accessed: 2025}
}

@article{sparckjones1972statistical,
  author  = {Spärck Jones, Karen},
  title   = {A Statistical Interpretation of Term Specificity and Its Application in Retrieval},
  journal = {Journal of Documentation},
  volume  = {28},
  number  = {1},
  pages   = {11--21},
  year    = {1972},
  doi     = {10.1108/eb026526}
}

@conference{tranicart,
author={Václav Tran and Jakub Šmíd and Jiří Martínek and Ladislav Lenc and Pavel Král},
title={Large Language Models for Summarizing Czech Historical Documents and Beyond},
booktitle={Proceedings of the 17th International Conference on Agents and Artificial Intelligence - Volume 2: ICAART},
year={2025},
pages={798-804},
publisher={SciTePress},
organization={INSTICC},
doi={10.5220/0013374100003890},
isbn={978-989-758-731-3},
issn={2184-433X}
}

@inproceedings{Smirnov2022ComparativeAO,
  title={Comparative Analysis of Neural Translation Models based on Transformers Architecture},
  author={Alexander V. Smirnov and Nikolay Teslya and Nikolay Shilov and Diethard Frank and Elena Minina and Martin Kovacs},
  booktitle={International Conference on Enterprise Information Systems},
  year={2022},
  url={https://api.semanticscholar.org/CorpusID:248702486}
}

@article{low_resource_survey,
author = {Ranathunga, Surangika and Lee, En-Shiun Annie and Prifti Skenduli, Marjana and Shekhar, Ravi and Alam, Mehreen and Kaur, Rishemjit},
title = {Neural Machine Translation for Low-resource Languages: A Survey},
year = {2023},
issue_date = {November 2023},
publisher = {Association for Computing Machinery},
address = {New York, NY, USA},
volume = {55},
number = {11},
issn = {0360-0300},
url = {https://doi.org/10.1145/3567592},
doi = {10.1145/3567592},
journal = {ACM Comput. Surv.},
month = feb,
articleno = {229},
numpages = {37}
}

@article{kldivergence,
 ISSN = {00034851},
 URL = {http://www.jstor.org/stable/2236703},
 author = {S. Kullback and R. A. Leibler},
 journal = {The Annals of Mathematical Statistics},
 number = {1},
 pages = {79--86},
 publisher = {Institute of Mathematical Statistics},
 title = {On Information and Sufficiency},
 urldate = {2026-04-22},
 volume = {22},
 year = {1951}
}

\end{document}